\documentclass[letterpaper]{article} % DO NOT CHANGE THIS
\usepackage[arxiv]{aaai2026}  %subb
\usepackage{times}  % DO NOT CHANGE THIS
\usepackage{helvet}  % DO NOT CHANGE THIS
\usepackage{courier}  % DO NOT CHANGE THIS
\usepackage[hyphens]{url}  % DO NOT CHANGE THIS
\usepackage{graphicx} % DO NOT CHANGE THIS
\usepackage{natbib}  % DO NOT CHANGE THIS AND DO NOT ADD ANY OPTIONS TO IT
\usepackage{caption} % DO NOT CHANGE THIS AND DO NOT ADD ANY OPTIONS TO IT
\usepackage{xcolor}
\usepackage{subcaption}
\usepackage{booktabs}
\usepackage{array}
\usepackage{siunitx}
\usepackage{amsmath}
\usepackage{amssymb}
\usepackage{booktabs}
\usepackage{multirow}
\usepackage{colortbl} % rowcolor
\usepackage{multicol}
\usepackage{float}
\usepackage{algorithm}
\usepackage{algorithmic}

\usepackage{newfloat}
\usepackage{listings}

\usepackage{pifont} % Required for \ding{41}
\makeatletter
\newcommand*{\myfnsymbol}[1]{%
  \ensuremath{%
    \ifcase#1\or
     *
     \or
      \text{\ding{41}}\or
      1\or
      2\or
      3\or
      4\or
      5\or
      6\or
      7\else
      8\fi
  }%
}
\def\@fnsymbol#1{\myfnsymbol{#1}}
\def\thempfootnote{\myfnsymbol{\c@mpfootnote}}
\makeatother

\DeclareCaptionStyle{ruled}{labelfont=normalfont,labelsep=colon,strut=off} % DO NOT CHANGE THIS
\floatstyle{ruled}
\newfloat{listing}{tb}{lst}{}
\floatname{listing}{Listing}

\definecolor{linkcolor}{rgb}{0,0,0} %for sub version

\title{AffordanceSAM: Segment Anything Once More in Affordance Grounding}

\author{
    \vspace{-24pt}\\
    Dengyang Jiang$^{2,3,1}$\thanks{Internship at SGIT AI Lab, State Grid Corporation of China. }$^\dagger$~
    Zanyi Wang$^{4,1*\dagger}$ ~
    Hengzhuang Li$^{5,1*}$ ~
    Sizhe Dang$^{1}$ ~   
    Teli Ma$^{3}$  \\[0.2mm]
    Wei Wei$^{2}$ ~
    Guang Dai$^{1}$ ~
    Lei Zhang$^{2}$ ~
    Harry Yang$^{3}$ ~
    Mengmeng Wang$^{6,1}$\thanks{Corresponding author. $^\dagger$Equal contribution.}
        \\[1.0mm]  
    \fontsize{10.4pt}{9.84pt}\selectfont
    $^{1}$SGIT AI Lab \hspace{4mm} 
    $^{2}$NWPU \hspace{4mm}
    $^{3}$HKUST \hspace{4mm}
    $^{4}$XJTU \hspace{4mm}
    $^{5}$HUST \hspace{4mm}
    $^{6}$ZJUT \hspace{4mm}
    }

\begin{document}
\maketitle
\begin{abstract}
Building a generalized affordance grounding model to identify actionable regions on objects is vital for real-world applications. Existing methods to train the model can be divided into weakly and fully supervised ways. However, the former method requires a complex training framework design and can not infer new actions without an auxiliary prior. While the latter often struggle with limited annotated data and components trained from scratch despite being simpler. This study focuses on fully supervised affordance grounding and overcomes its limitations by proposing AffordanceSAM, which extends SAM's generalization capacity in segmentation to affordance grounding. Specifically, we design an affordance-adaption module and curate a coarse-to-fine annotated dataset called C2F-Aff to thoroughly transfer SAM's robust performance to affordance in a three-stage training manner. Experimental results confirm that AffordanceSAM achieves state-
of-the-art (SOTA) performance on the AGD20K benchmark and exhibits strong generalized capacity.
\end{abstract}

\section{Introduction}
\label{sec:introduction}
Affordance grounding~\cite{aff1} refers to finding potential “action possibilities” regions of an object, which plays a key role in bridging the gap between visual perception and robotic action. Recently, attempts made to endow models to have such grounding abilities can be broadly divided into two ways. The first type of methods is weakly supervised affordance grounding~\cite{locate,PLSP,r-mamba}, which aims to identify affordance regions on objects using human-object interaction images and egocentric object images without dense labels. The affordance maps are obtained via class activation mapping (CAM)~\cite{cam} or an auxiliary prior model~\cite{sam1}. However, these methods require a complex training framework design since there are two branches (exocentric and egocentric) that need to be balanced and optimized during training. Moreover, the generalization performance of these models is suboptimal, as their supported classes are fixed (e.g, LOCATE~\cite{locate}) or very few parameters are optimized on the affordance task (e.g., PLSP~\cite{PLSP}). To tackle this problem, the second type of methods uses affordance maps to directly supervise the models~\cite{affordancellm,ooal}. However, some important affordance components trained from scratch (e.g., decoder) on only hundreds of pieces of manually labeled training data make it insufficient to obtain a highly generalized model. Thus, it is important for the fully supervised affordance grounding family to \textbf{find a strong foundation model that is naturally suitable for the affordance grounding task} and \textbf{scale up the supervised data} to obtain the ideal generalized model.

\begin{figure}[t]
    \centering
    \includegraphics[width=1\linewidth]{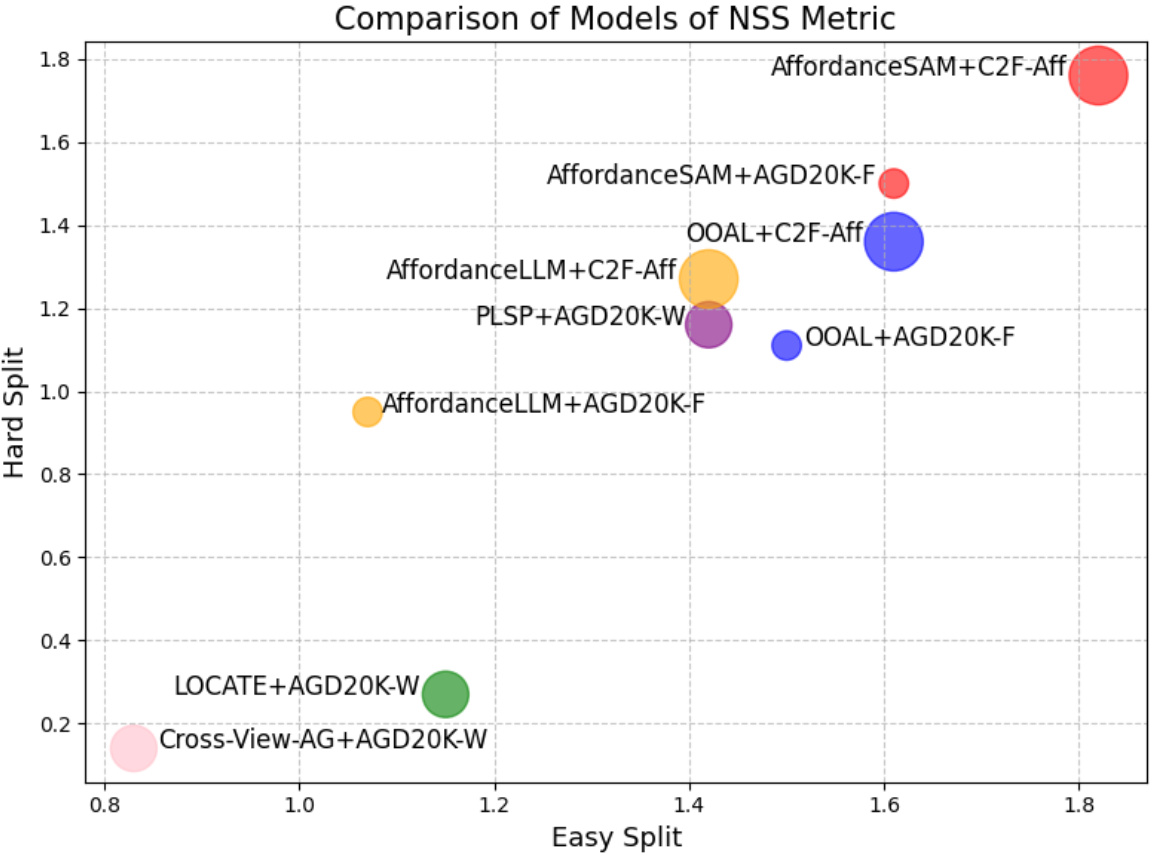}
    \caption{\textbf{Performance Comparison}: The circle area indicates the number of training data, with better-performing models positioned toward the upper right. Our AffordanceSAM and C2F-Aff data can respectively serve as an excellent base model and training data. Integrating the two achieves a performance far ahead of other candidates.}
    \vspace{-1em}
    \label{fig:begin_images}
\end{figure}
In this work, we try to decompose and address this problem in a step-by-step manner:

\textbf{(i)} \textit{Incorporating a suitable and generalized vision foundation model}. With the recent advancements in large-scale pre-trained foundation models, many works have attempted to leverage the prior knowledge for downstream task transfer~\cite{action-clip,fewshot-dino}. We believe this paradigm is also effective for the affordance grounding task. Given that affordance grounding inherently requires pixel-level localization, we explore the possibility of incorporating SAM (Segment Anything Model)~\cite{sam1,sam2} since SAM not only demonstrates remarkable generalization capabilities but also excels in precise object localization after pre-trained on large-scale masked-labeled data. Moreover, the model structure and decoder output format of segmentation and affordance grounding are highly similar. Considering these similarities between segmentation and affordance grounding, we hypothesize that incorporating SAM could be well-suited for our problem. 

\begin{figure}[h!]
    \centering
    \vspace{-1mm}
\includegraphics[width=0.88\linewidth]{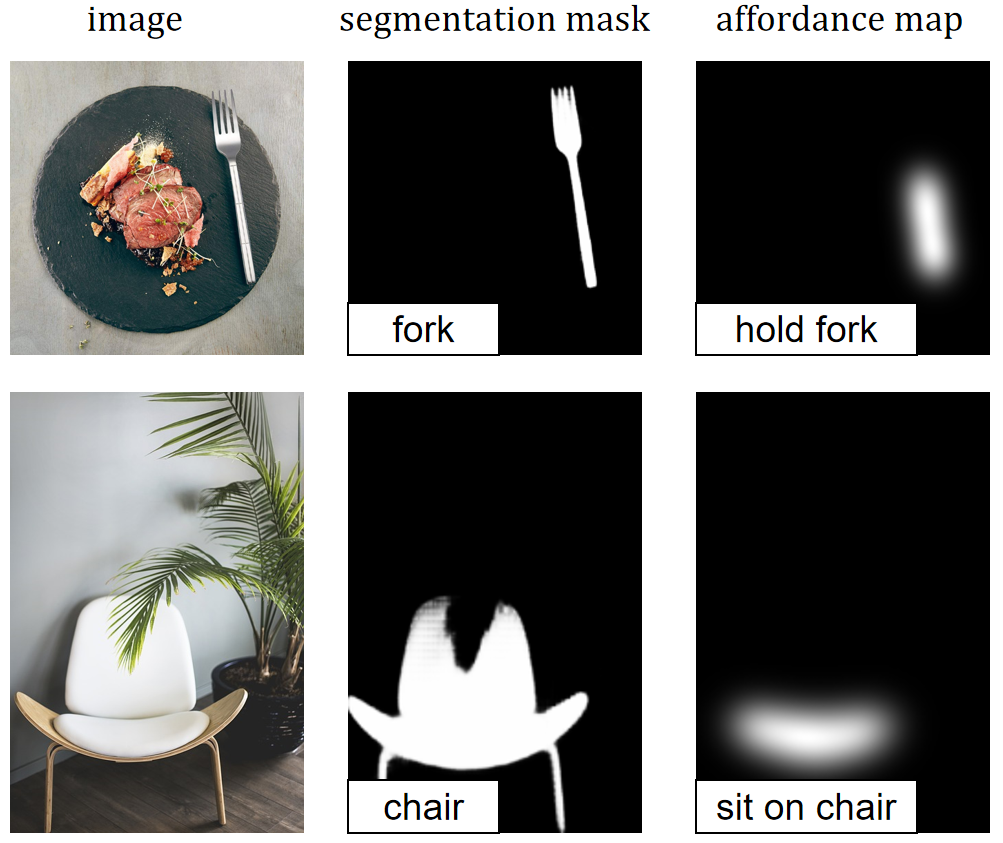}
    \caption{\textbf{Differences of two task,} where segmentation focuses on separating objects according to the prompt, but affordance emphasizes grounding the possible effective part of objects based on the affordance query.}
    \vspace{-0.5em}
    \label{fig:difference}
\end{figure}
Although there are some similarities between segmentation and affordance grounding, task differences still exist. As shown in Figure~\ref{fig:difference}, firstly, affordance grounding requires the recognition of functional parts of objects based on verb queries, while segmentation requires the separation of objects; the ground-truth of segmentation and affordance also differs slightly: the former uses a binary mask, while the latter uses a heatmap that encodes functional possibilities.

\textbf{(ii)} \textit{Effectively transferring the capabilities of SAM through data and model structure aspects}. As mentioned above, there exist some differences of these two tasks, to fully unlock SAM's capacity for affordance grounding, we first design an affordance-adaption module that utilizes learnable queries to interact with text and image features. These queries are then sent into the decoder to refine the original mask produced by SAM. We hope that this prompt learning-like module can effectively search for affordance-relevant parts in the comprehensive knowledge of SAM. We then scale up the fully supervised data in a coarse-to-fine annotated manner, which we named C2F-Aff. The dataset consists of three parts with each corresponding to a specific label format, and thus the training process is divided into three stages. In the first data part and training stage, we curate a mask-labeled dataset by merging and reclassifying existing datasets, focusing on object-affordance pairings. This enables the model to learn basic object-verb relationships. After that, we use a weakly supervised model to generate pseudo-labels for unannotated images in AGD20K~\cite{AGD20K} and post-process them to obtain annotations with higher quality to finetune the model. Finally, we use the high-quality, human-labeled data
in AGD20K for further fine-tuning the model to boost the performance.

In our comprehensive experiments, we show that AffordanceSAM and C2F-Aff data can severally serve as an excellent base model and training data, which is evidenced by the leading performance of AffordanceSAM against other methods using the same amount of data and the performance enhancement of a wide range of fully supervised methods once trained on our C2F-Aff data.

In summary, our contributions are as follows:
\begin{itemize}
\item Showing that SAM is natural fit for affordance grounding with analysis and proposed AffordanceSAM.
\item Introducing C2F-Aff dataset and a three-stage training scheme that boosts the performance of both AffordanceSAM and prior fully-supervised methods. 
\item Achieving SOTA performance on the AGD20K benchmark, as well as showing the evidence to generalize to novel actions and objects.
\end{itemize}

\section{Related Work} 
We discuss the most relevant studies here and provide a more discussion in Appendix \textcolor{linkcolor}{1}.

\noindent\textbf{Affordance Grounding.} Unlike detection~\cite{detect} and segmentation~\cite{seg} that yield bounding boxes or binary masks of the desired objects, affordance grounding~\cite{aff1} needs a model to output the functional possibilities in a object. 
Early works~\cite{hotspot,eil,spa} have tried to equip the model with such capacity. But because of the data restriction at that time, these models can only recognize a few types of objects and actions. Next, Luo et al. collect the first large-scale affordance dataset called AGD20K~\cite{AGD20K} and annotated a few of the data with affordance maps (1675/23816). After that, many works engined by AGD20K have emerged. For example, LOCATE~\cite{locate} localizes and extracts affordance-specific information in exocentric and transfers this knowledge to egocentric in a weakly supervised manner; AffordanceLLM~\cite{affordancellm} builds a vision-language learning framework based on LLaVA~\cite{llava} to conduct the open affordance learning. PLSP~\cite{PLSP} uses an additional semantic prior (e.g., SAM~\cite{sam1})\footnote{While both PLSP and our work use SAM, PLSP only uses SAM as a prior and refiner to guide and assist the main model branch. By contrast, our method directly tunes SAM to be an affordance grounding model in a fully supervised manner.} to guide the traditional two-branch weakly supervised training process.
However, none of the above works show satisfactory generalization ability to recognize regions for unseen objects and affordance actions (reasons are explained in the Introduction, and evidence is shown in our experiment results). In this study, we aim to find a foundation model
that is naturally suitable for the task and scale up the annotated data to obtain such an ideal model.

\begin{figure*}[t]
\centering
\vspace{-1.5em}
\begin{subfigure}[t]{0.552\textwidth}
    \centering
    \includegraphics[width=\textwidth]{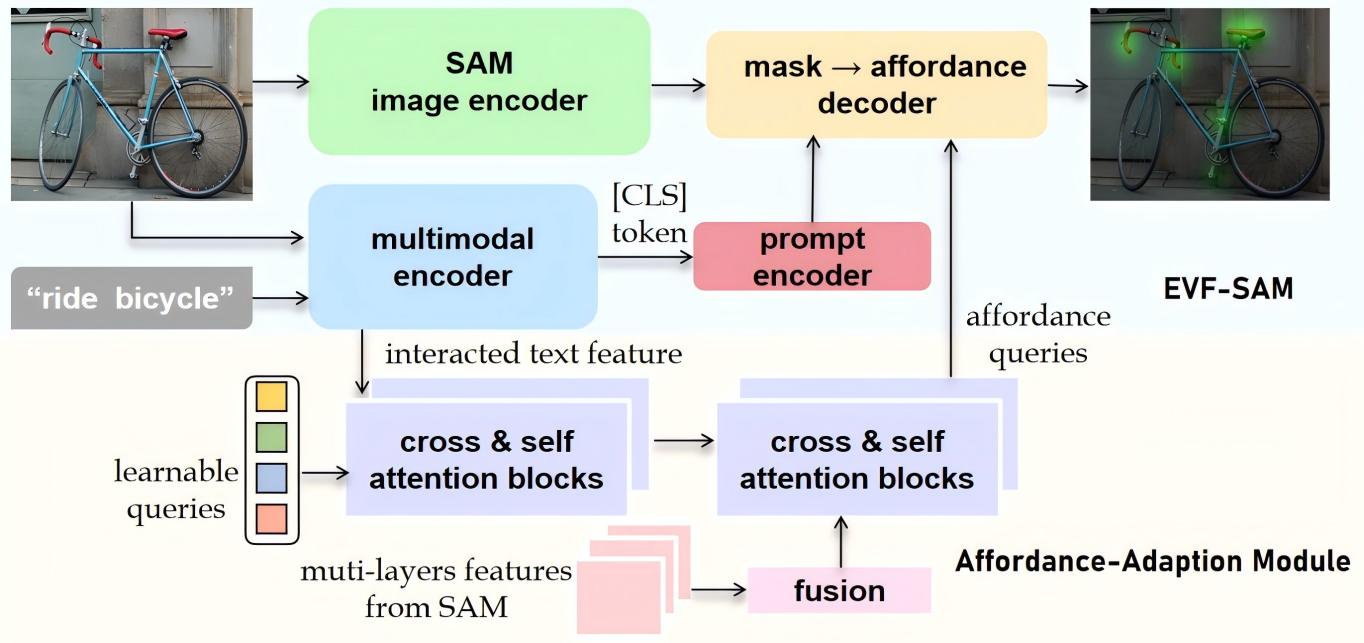}
    \caption{\textbf{Overview of AffordanceSAM architecture.}}
    \label{fig:arch}
\end{subfigure}
\hfill
\begin{subfigure}[t]{0.435\textwidth}
    \centering
    \includegraphics[width=\textwidth]{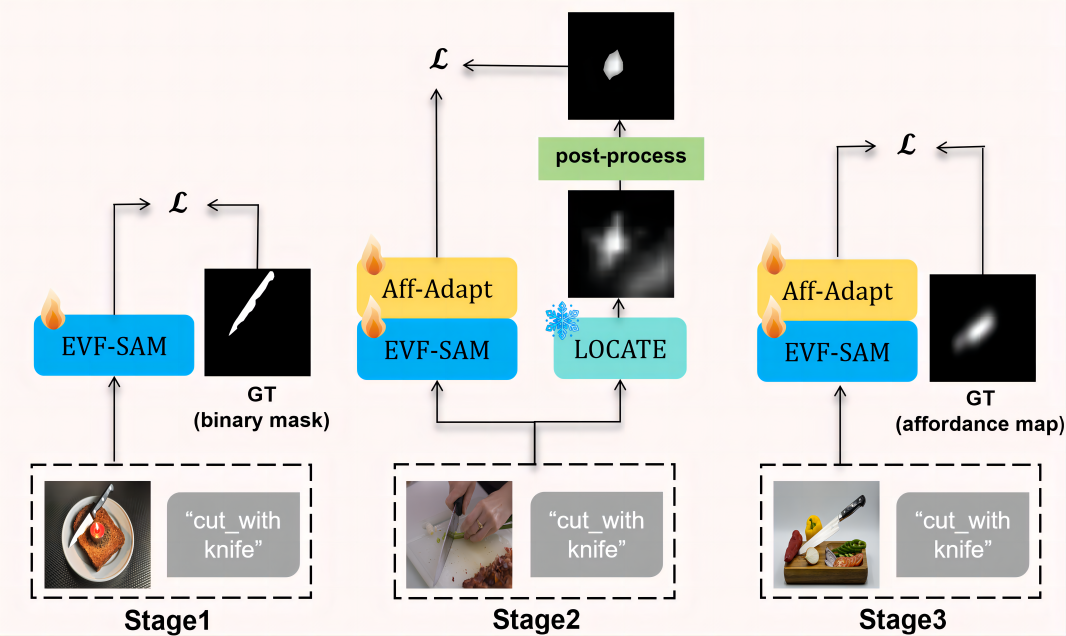}
    \caption{\textbf{Overview of AffordanceSAM training recipe.}}
    \label{fig:training}
\end{subfigure}
  \caption{\textbf{(a). Architecture:} AffordanceSAM is built upon EVF-SAM ~\cite{evfsam} with an additional affordance-adaption module.
  % The inputs of our model include a single image and a text prompt related to interaction. In our affordance-adaption module, we employ learnable queries to fuse with text feature and multiple layers of features from SAM encoder and then incorporate them into the original decoder to help the transformation from mask to the final affordance map. 
  \textbf{(b). Training:} Trained on purposed C2F-Aff dataset, AffordanceSAM adopts a coarse-to-fine training recipe. 
  %In stage1 and stage3, we use ground truth binary masks and affordance maps to supervise the model, respectively. In stage2, affordance maps output by LOCATE and then pass through our proposed post-process program are used to supervise the model.
  }
 \vspace{-0.5em}
\end{figure*}
\noindent\textbf{Foundation Model for Downstream Tasks.}
Vision foundation models (e.g., SAM~\cite{sam1}, CLIP~\cite{clip}, DINOv2~\cite{dinov2}) are large-sized models trained over vast amounts of data . After training, they are suitable as a starting point for a variety of downstream tasks under limited data conditions. Recently, many works have used different foundation models for different downstream tasks~\cite{motiondirector,action-clip,fewshot-dino,promptlearn3}. For example, ActionCLIP~\cite{action-clip} proposes a “pre-train, prompt and fine-tune” strategy to transfer CLIP to action recognition;  FM-FSOD~\cite{fewshot-dino} uses DINOv2 to extract fine-grained features for few-shot object detection. Our work also shares similarities, while we focus on adapting SAM to the affordance grounding task with proposed affordance-adaption module and curated data.

\section{Approach}
\label{sec:approach}
The overall model architecture and training procedure of AffordanceSAM are shown in Figure~\ref{fig:arch} and Figure~\ref{fig:training}. As our model consists of two primary components: the EVF-SAM~\cite{evfsam}\footnote{EVF-SAM can be seen as an 
extension version of SAM that support text prompts. In this paper, we use term `EVF-SAM' to refer the specific model part and `SAM' for our general argument.} baseline and the affordance-adaption module, we begin with describing each, and finaly, we provide a comprehensive elucidation of our coarse-to-fine affordance (C2F-Aff) dataset and training recipe.
\subsection{Preliminary}
\label{app:pre}
\begin{table*}[t]
    \centering
    \vspace{-1.5em}
    \resizebox{0.78\linewidth}{!}{
    \begin{tabular}{c|c|c|c}
    \toprule
        Part & Data Source & Label Type & Num. \\
    \midrule
\multirow{2}{*}{1} & \multirow{2}{*}{PADv2 / Handal / RGB-D Part Affordance} & \multirow{2}{*}{Binary Mask} & \multirow{2}{*}{39159} \\
& & & \\
\midrule
\multirow{2}{*}{2} & \multirow{2}{*}{Unlabeled Part of AGD20K} & \multirow{2}{*}{Pseudo Labeled Affordance Map} & 13323 (Easy Split) \\
& & & 11889 (Hard Split) \\
\midrule
\multirow{2}{*}{3} & \multirow{2}{*}{Labeled Part of AGD20K} & \multirow{2}{*}{Human Annotated Affordance Map} & 1135 (Easy Split) \\
& & & 868 (Hard Split) \\
    \bottomrule
    \end{tabular}}
    \caption{\textbf{Overview of the training dataset C2F-Aff.} This table summarizes the data source, format of groundtruth, the number of samples for training in each stage. More details can be found in Appendix~\textcolor{linkcolor}{2}.}
    \label{tab:datasets}
    \vspace{-0.5em}
\end{table*}
\noindent\textbf{Baseline EVF-SAM.} As we need to provide text prompts that contain affordance queries to the model but the original SAM lacks language understanding abilities. We choose to start from EVF-SAM~\cite{evfsam} which is the SOTA method among attempts~\cite{evfsam,lisa,groundedsam,fastsam} that empowered SAM to follow text instructions. We seek to leverage its robust segmentation capabilities for affordance grounding.

The EVF-SAM original framework incorporates four key components: a multimodal encoder (BEIT-3~\cite{beit3}) denoted as $\mathcal{E}_M$, a prompt encoder $\mathcal{E}_P$, a SAM image encoder $\mathcal{E}_I$, and a SAM mask decoder $\mathcal{D}$. The input image is firstly processed through two parallel paths: (i): conversion to SAM image tokens $\mathbf{I}_s \in \mathbb{R}^{B \times N_s \times D_s}$, and (ii): transformation to multimodal image tokens $\mathbf{I}_m \in \mathbb{R}^{B \times N_m \times D_m}$, where $B$ represents batch size, $N$ sequence length, and $D$ feature dimension. Meanwhile, the text input is first tokenized to produce text tokens $\mathbf{T} \in \mathbb{R}^{B \times N_t \times D_m}$. These text and multimodal image tokens are then combined with a learnable {\tt[CLS]} token $\in \mathbb{R}^{B \times 1 \times D_m}$ and processed by the multimodal encoder:
\begin{equation}
\mathbf{F}_m = \mathcal{E}_M([\texttt{[CLS]}; \mathbf{I}_m; \mathbf{T}]) \in \mathbb{R}^{B \times (1 + N_m + N_t) \times D_m}.
\end{equation}
The {\tt[CLS]} token output $\mathbf{F}_c$ which is split from $\mathbf{F}_m$ is transformed by the prompt encoder, then together with the SAM encoded image features to generate binary mask $\mathbf{M}_b$:
\begin{equation}
\mathbf{M}_b = \mathcal{D}(\mathcal{E}_I(\mathbf{I}_s), \mathcal{E}_P(\mathbf{F}_c)).
\end{equation}

\subsection{Affordance-Adaption Module}
\label{app:affcon}
As shown in Figure~\ref{fig:difference}, there exists a substantial disparity between the output of the segmentation task and the affordance grounding task. To efficiently adapt SAM to the dataset labeled with affordance maps instead of its original output masks, we introduce this module.
 Similar to BLIP-2~\cite{blip2} and DETR~\cite{detr}, we introduce a set of learnable queries termed affordance queries. These queries, denoted as $\mathbf{Q}_a\in {N_s \times D_m}$, first repeat over batch dimension and conduct cross-attention with the text features that interact with the image in the multimodal encoder. we expect these queries to extract information concerning the affordance action from these semantic features. What's more, since different layers of SAM's features often exhibit different levels of granularity and knowledge~\cite{hqsam}, and affordance may correspond to multiple parts of an object. We consider a diverse set of granularities can be beneficial.  Therefore, the second cross-attention operation is performed between the processed affordance queries and fused visual features of SAM's multiple layers formulated as:
\begin{equation}
\mathbf{F}v = \sum_{i=1}^{j} \alpha_i \cdot Linear(\mathbf{G}_i), \quad \alpha_1 + \ldots + \alpha_j = 1,
\end{equation}
where $\mathbf{G}_i\in \mathbb{R}^{B \times N_s \times D_s}$ denotes features from the global-attention blocks in SAM image encoder, $\alpha_i$ is a learnable parameter that controls the fusion ratio of each feature. It is worth noting that we also add one self-attention block after each cross-attention block for enhanced feature learning. The final affordance queries $\mathbf{Q}_{af}$ are first processed by two transposed convolution functions to adjust the dimension, then added with mask features from the original EVF-SAM to obtain the modified features for final process. Thus, we can obtain the affordance map output $\mathbf{M}_a$ as:
\begin{equation}
\mathbf{M}_a = \mathcal{D}(\mathcal{E}_I(\mathbf{I}_s), \mathcal{E}_P(\mathbf{F}_c), TransConv(\mathbf{Q}_{af})).
\end{equation}

\subsection{Coarse-to-Fine Affordance Dataset and Training}
\label{app:training}
To fully leverage SAM's generalization capacity for affordance grounding, we craft a coarse-to-fine training dataset (C2F-Aff) and training recipe shown in Figure~\ref{fig:training}. The brief summary of C2F-Aff is displayed in Table~\ref{tab:datasets}, and more information on our data collection, division, organization and cleaning can be found in Appendix \textcolor{linkcolor}{2}. We divide the data and training into three parts based on the form of the labels, while we always use a template of ``\texttt{<affordance action>} \texttt{<object\_name>}'', for instance, ``\texttt{wear} \texttt{hat}'', as the text prompt for AffordanceSAM. 
%The detailed dataset setup are provided in Appendix~\textcolor{red}{2}.

\noindent\textbf{Dataset Part and Training Stage 1.} In most training datasets for multimodal models and text-prompted segmentation models, such as LAION-400M~\cite{laion}, RefCLEF~\cite{refclff}, and RefCOCO~\cite{refcoco}, there are very few objects with affordance properties and very little textual content that contains verbs. However, when transferring a foundation model to affordance grounding task in a fully supervised manner (e.g, DINOv2 in OOAL, LLaVA in AffordanceLLM, and SAM in our method), it is crucial for them to familiarize themselves with such objects and understand the affordance verb information for locating potential “action-possibility” regions. Thus, In the first part of our C2F-Aff dataset, we select three datasets, PADv2~\cite{pad}, Handal~\cite{handal}, and RGB-D Part Affordance~\cite{partaff}, which satisfy both objects contained with affordance actions and annotations of masks. We then combine them according to the categories of objects and affordance actions (details are in Appendix \textcolor{linkcolor}{2}). 
And for We only use EVF-SAM baseline without the affordance-adaption module and fine-tune multimodal encoder and prompt encoder in this stage. The training loss function is given by
\begin{equation}
    \mathcal{L} = \lambda_{dice} \cdot \mathcal{L}_{dice} + \lambda_{bce} \cdot \mathcal{L}_{bce},
\end{equation}
where the overall loss  $\mathcal{L}$ is the weighted sum of DICE loss  $\mathcal{L}_{dice}$~\cite{dice} and BCE loss  $\mathcal{L}_{bce}$~\cite{bce}, determined by $\lambda_{dice}$ and $\lambda_{bce}$.

\noindent\textbf{Dataset Part and Training Stage 2.} As mentioned in the Introduction, images with dense pixel labeling affordance maps are scarce and it can be costly for us to annotate. In practice, we find that training only on the hundreds of labeled images could not fully unlock SAM's capacity. Hence, we utilize a weakly supervised model on AGD20K, LOCATE~\cite{locate}, to label the vast majority of the remaining unlabeled images in AGD20K. However, The raw output affordance heatmaps of LOCATE are usually suboptimal, which exist a large area of low thermal regions unrelated to the target. Hence, we design a post-processing program showed in Algorithm~\ref{alg:process} to mitigate such issue. These processed pseudo labels help bridge the gap between binary segmentation masks from part 1 and the detailed affordance heatmaps needed for the final output, serving as an intermediate step in our C2F-Aff dataset. 
During training, we add affordance-adaption module and keep only SAM image encoder frozen. we follow AffordanceLLM~\cite{affordancellm} to use binary focal loss~\cite{focal} ($\mathcal{L}_{focal}$) and set the weight of positive examples to be 0.9 and that of negative ones to be 0.1, as there are often more negatives than positives in an affordance map.
\begin{algorithm}[h!]
\caption{Affordance maps post-process algorithm.}
\textbf{Input}: Affordance map $M_{al}$ generated by LOCATE and threshold $\gamma$.
\begin{algorithmic}[1]
\STATE $\gamma_1 = \gamma \cdot max(M_{al})$
\STATE $\gamma_2 = 0.4 \cdot \gamma_1$
\STATE $\gamma_3 = 0.2 \cdot \gamma_2$
\STATE $M_{ap} = where(M_{al} \ge \gamma_1, M_{al} , \frac{{M_{al}}^2}{\gamma_1})$
\STATE $M_{ap} = where(M_{ap} \ge \gamma_2, M_{ap} , \frac{{M_{ap}}^2}{\gamma_2})$
\STATE $M_{ap} = where(M_{ap} \ge \gamma_3, M_{ap} , 0)$
\end{algorithmic}
\label{alg:process}
\textbf{Output}: Processed affordance map $M_{ap}$.
\end{algorithm}

\noindent\textbf{Dataset Part and Training Stage 3.} To achieve the final high-quality supervised fine-tuning for boosting performance, we use the high-quality human-labeled samples in AGD20K as the third part of our C2F-Aff dataset and use them to supervise the model. We aim at regulating the model to generate the most precise and fine-grained affordance maps. We keep the loss function and trainable modules unchanged compared to stage 2.

\begin{table*}[t]
\centering
\vspace{-1.0em}
\caption{\textbf{Quantitative results on AGD20K benchmark.} The \textbf{best} and \underline{second-best} results are marked in bold and underlined. }
\vspace{-0.5em}
\resizebox{0.85\linewidth}{!}{
\begin{tabular}{llccc|ccc}
\toprule
\multirow{2}{*}{Method} & \multirow{2}{*}{Venue} & \multicolumn{3}{c|}{Easy Split} & \multicolumn{3}{c}{Hard Split} \\
\cmidrule(lr){3-5} \cmidrule(l){6-8}
& & {KLD$\downarrow$} & {SIM$\uparrow$} & {NSS$\uparrow$} & {KLD$\downarrow$} & {SIM$\uparrow$} & {NSS$\uparrow$} \\
\midrule
\multicolumn{8}{l}{\textit{Weakly supervised methods trained on AGD20K-Weakly ($\sim$12K) }} \\
Cross-View-AG~\cite{AGD20K} & CVPR'22 & 1.787 & 0.285 & 0.829 & 2.092 & 0.209 & 0.138 \\
LOCATE~\cite{locate} & CVPR'23 & 1.405 & 0.372 & 1.157 & 1.829 & 0.282 & 0.276 \\
R-Mamba~\cite{r-mamba} & CVPR'25 & 1.310 & 0.397 & 1.279 & - & - & - \\
PLSP~\cite{PLSP} & ICLR'25 & 1.153 & 0.437 & 1.418 & 1.401 & 0.395 & 1.109 \\
\midrule
\multicolumn{8}{l}{\textit{Fully supervised methods trained on AGD20K-Fully ($\sim$1K)}} \\
AffordanceLLM~\cite{affordancellm} & CVPR'24 & 1.463 & 0.377 & 1.070 & 1.661 & 0.361 & 0.947 \\
OOAL~\cite{ooal} & CVPR'24 & \underline{1.070} & 0.461 & 1.503 & 1.302 & 0.410 & 1.119\\
AffordanceSAM & this work & 1.271 & 0.486 & 1.597 & 1.327 & 0.423 & \underline{1.502} \\

\midrule
\multicolumn{8}{l}{\textit{Fully supervised methods trained on C2F-Aff ($\sim$40K + $\sim$12K + $\sim$1K)}} \\
AffordanceLLM~\cite{affordancellm} & CVPR'24 & 1.170 & 0.482 & 1.425 & 1.312 & 0.405 & 1.293 \\
OOAL~\cite{ooal} & CVPR'24 & \textbf{0.974} & \underline{0.504} & \underline{1.650} & \textbf{1.119} & \underline{0.442} & 1.364\\
AffordanceSAM & this work & 1.083 & \textbf{0.543} & \textbf{1.800} & \underline{1.128} & \textbf{0.514} & \textbf{1.761} \\

\bottomrule
\end{tabular}}
\label{tab:main}
\vspace{-0.5em}
\end{table*}
\begin{figure*}[h!]
    \centering
    \includegraphics[width=0.80\linewidth]{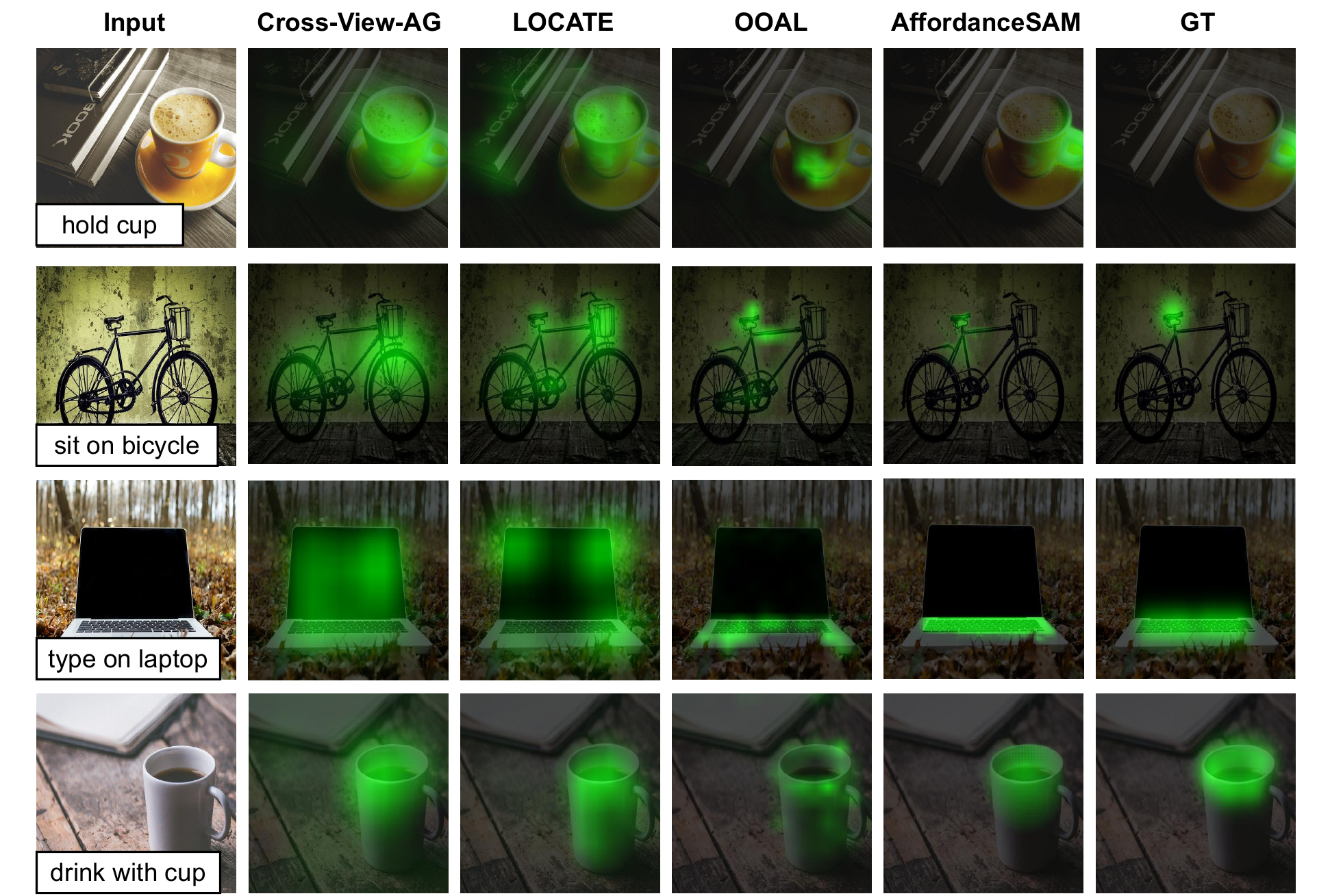}
    \caption{\textbf{Qualitative comparison on AGD20K dataset.} Cross-View-AG and LOCATE often make affordance predictions with a large area of low thermal regions unrelated to the target. On the contrary, OOAL focuses on a small area but sometimes it's completely wrong. Our AffordanceSAM can precisely recognize the part of the object that contains the affordance action.}
    \vspace{-0.5em}
    \label{fig:agd}
\end{figure*}
\begin{figure*}[t]
    \vspace{-1.5em}
    \centering
    \includegraphics[width=1\linewidth]{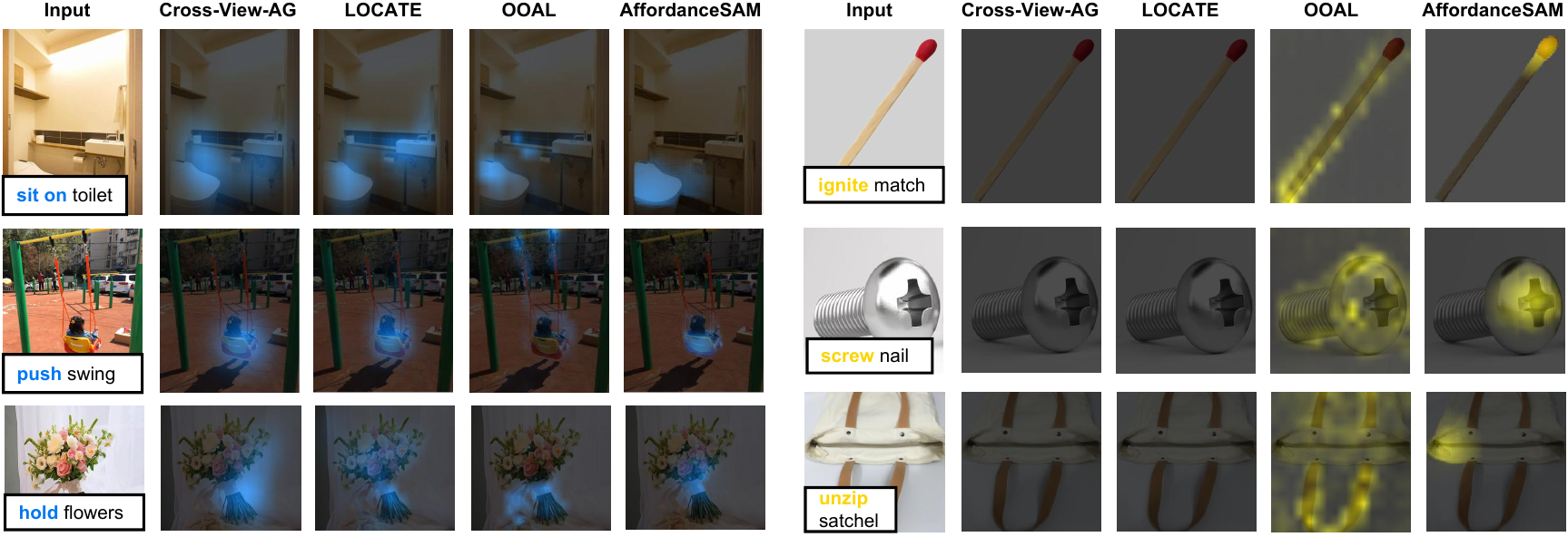}
    \caption{\textbf{Qualitative results with novel objects from the internet.} \textcolor[RGB]{0,127,254}{learned} and \textcolor[RGB]{251,213,3}{novel} affordance actions are marked in blue and yellow. Cross-View-AG and LOCATE can not generalize to new affordance actions, so we do not highlight any region. OOAL often outputs suboptimal affordance maps when encountering new objects and affordance actions. By contrast, our AffordanceSAM can still give reasonable affordance maps.}
    \vspace{-0.5em}
    \label{fig:internet}
\end{figure*}
\section{Experiments}
\label{sec:experiments}
\subsection{Experimental Setup}

\noindent\textbf{Implementation Details.} In stage1, we initialize our AffordanceSAM with EVF-SAM~\cite{evfsam} and train for 13 epochs with base learning as 0.00002. In stage2, we use the weights obtained from the first stage and randomly initialize affordance-adaption module, we train the model with the same number of epochs and base learning as stage1. In stage3, we further train our model for 26 epochs with larger base learning as 0.00004. In all training stages, we employ an AdamW optimizer~\cite{adam} with a cosine learning rate scheduler, and the total batchsize per iteration is 32 using 4 NVIDIA A100 (80GB) GPUs.

\noindent\textbf{Evaluation.}
 We strictly follow the settings in AffordanceLLM to have two splits (easy split and hard split) of AGD20K to test our model. The easy split is the original unseen split of AGD20K, which has a lot of similarities between the objects in the train
and test sets. The hard split ensures that there is no overlap between the object categories in the train and test set (both in our stage 2 and stage 3). Meanwhile, we follow the metrics provided in AGD20K to evaluate our model, which is Kullback-Leibler Divergence (KLD)~\cite{kld}, Similarity (SIM)~\cite{sim} and Normalized Scanpath Saliency (NSS)~\cite{nss}.

\noindent\textbf{Methods for Comparison.} 
Affordance grounding methods can be mainly summarized into two main categories: weakly supervised and fully supervised methods. We compare our approach against representative weakly supervised baselines (Cross-View-AG~\cite{AGD20K}, LOCATE~\cite{locate}, R-Mamba~\cite{r-mamba}, and PLSP~\cite{PLSP}), and  fully supervised baselines (AffordanceLLM~\cite{affordancellm} and OOAL~\cite{ooal}). 

We also provide more details about our implementation including training hyperparameters and model configuration, details of each metric, and detailed descriptions of each baseline method for comparison in Appendix~ \textcolor{linkcolor}{3$\sim$5}.

\subsection{Main Results}

We present both quantitative and qualitative experimental results and analyses of our proposed method. Below, we summarize the key findings:

\noindent\textbf{Effectiveness of AffordanceSAM Model and C2F-Aff Dataset.}
As illustrated in Table~\ref{tab:main}, the AffordanceSAM model demonstrates remarkable performance, achieving competitive results even without leveraging a larger dataset (C2F-Aff). This is particularly notable when compared to weakly supervised methods. Trained only the AGD20K-Fully, AffordanceSAM outperforms AffordanceLLM and outperforms the previous SOTA OOAL model in terms of KLD and NSS, across both the easy and hard splits. From a data perspective, the introduction of the C2F-Aff dataset has significantly enhanced the performance of all fully supervised methods in the challenging hard splits, and the improvement is most prominent for our model. It is worth noting that all the fully supervised methods trained on  C2F-Aff dataset outperformed other weakly supervised candidates. This illustrates the potential of the fully supervised method when scaling the supervised data.

\noindent\textbf{Generalized Performance when Combining the Model and Dataset.}
By combining the model with the rich annotations of C2F-Aff, AffordanceSAM achieves state-of-the-art (SOTA) performance across most metrics. Specifically, AffordanceSAM attains the highest SIM and NSS scores on the easy and hard splits, respectively. Furthermore, AffordanceSAM demonstrates exceptional generalization to novel objects and actions, as evidenced by qualitative results on both AGD20K images and internet images (Figure~\ref{fig:agd} and Figure~\ref{fig:internet}). For instance, when handling the object ``\texttt{cup}", AffordanceSAM demonstrates a superior understanding of affordance actions compared to other methods and generates more accurate and precise affordance maps according to different actions ``\texttt{hold}" and ``\texttt{drink}". These results substantiate that our AffordanceSAM can benefit so greatly from the foundation model SAM, as well as the proposed C2F-Aff dataset, thus showcasing leading affordance grounding performance towards novel objects and affordance actions.

\subsection{Ablation Study}
In this section, we conduct an extensive ablation study to reveal the contribution and design-space of each component. Unless otherwise specified, we report the results on hard split of AffordanceSAM using our default settings. Please refer to Appendix~\textcolor{linkcolor}{6} to see the detailed illustration of the evaluation setup of each table. 
\begin{table*}[t]
\centering
\vspace{-1em}
% First row with two tables side by side
\begin{minipage}[t]{0.6\textwidth}
\centering
\begin{tabular}{lllccc}
\hline
Part 1 & Part 2 & Part 3 & KLD $\downarrow$ & SIM $\uparrow$ & NSS $\uparrow$ \\
\hline
$\times$ & $\times$ & $\times$ & 2.924 & 0.184 & 0.115 \\
$\checkmark$ & $\times$ & $\times$ & 2.280 & 0.289 & 0.741\\
$\checkmark$ & $\checkmark$ & $\times$ & 1.592 & 0.396 & 1.319 \\
\rowcolor{gray!16}
$\checkmark$ & $\checkmark$ & $\checkmark$  & \textbf{1.128} & \textbf{0.514} & \textbf{1.761} \\
\multicolumn{3}{l}{\emph{Combining all the data}} & {1.735} & {0.361} & {1.265} \\
\hline
\end{tabular}
\caption{\textbf{Ablation results of coarse-to-fine dataset and training recipe.} First, we gradually add the training data of each part in C2F-Aff . Then, we provide the results when directly combine all the data and train for one stage.}
\label{subtab:training_reciep}
\end{minipage}
\hfill
\begin{minipage}[t]{0.35\textwidth}
\centering
\begin{tabular}{lccc}
\hline
Num. & KLD $\downarrow$ & SIM $\uparrow$ & NSS $\uparrow$ \\
\hline
1 & 1.408 & 0.358 & 1.562 \\
2 & 1.198 & 0.503 & 1.700 \\
\rowcolor{gray!16}
3 & \textbf{1.128} & \textbf{0.514} & \textbf{1.761}  \\
\hline
\end{tabular}
\setcounter{table}{4}
\caption{\textbf{Ablation results of number of filtration.} Num. indicates  the number of times we filter the affordance maps shown in steps 4$\sim$6 of Algorithm~\ref{alg:process}.}
\addtocounter{table}{-1}
\label{subtab:fli_num}
\end{minipage}

\vspace{1em}

% Second row with two tables side by side
\begin{minipage}[t]{0.54\textwidth}
\centering
\begin{tabular}{lccc}
\hline
Modules & KLD $\downarrow$ & SIM $\uparrow$ & NSS $\uparrow$ \\
\hline
EVF-SAM only & 1.327 & 0.422 & 1.615 \\
w/ $LQ$ & 1.221 & 0.475 & 1.695 \\
w/ $LQ, F_t$ & 1.218 & 0.487 & 1.703 \\
\rowcolor{gray!16}
w/ $LQ, F_t, F_v \text{ w/ m}$ & \textbf{1.128} & \textbf{0.514} & \textbf{1.761} \\
w/ $LQ, F_t, F_v \text{ wo/ m}$ & 1.192 & 0.503 & 1.749 \\
\hline
\end{tabular}
 \setcounter{table}{3}
\caption{\textbf{Ablation results of affordance-adaption module.} $LQ$: learnable queries. $F_t$: interaction with text feature. $F_v \text{ w/ m}$: interaction with  fused image features. $F_v \text{ w/o m}$: interaction with only last layer image feature.}
\addtocounter{table}{1}
\label{subtab:affcon}
\end{minipage}%
\hfill
\begin{minipage}[t]{0.41\textwidth}
\centering
\begin{tabular}{lccc}
\hline
$\gamma$ & KLD $\downarrow$ & SIM $\uparrow$ & NSS $\uparrow$ \\
\hline
0 & 1.394 & 0.387 & 1.693 \\
0.3 & 1.231 & 0.465 & 1.679 \\
\rowcolor{gray!16}
0.45 & \textbf{1.128} & \textbf{0.514} & \textbf{1.761}  \\
0.6 & 1.202 & 0.474 & 1.665 \\
0.9 & 1.360 & 0.410 & 1.497 \\
\hline
\end{tabular}
\caption{\textbf{Ablation results of $\gamma$.} $\gamma$ indicates the intensity of filtration when we post-process affordance maps produced by LOCATE.}
\label{subtab:postprocess}
\end{minipage}

\caption{\textbf{AffordanceSAM ablation experiments on the hard split.} The \colorbox{gray!16}{default setting} is marked in gray.}
\label{tab:ablation}
\vspace{-2mm}
\end{table*}

\noindent\textbf{Contribution of Coarse-to-Fine Dataset and Training Recipe.} As depicted in Table~\ref{subtab:training_reciep}, our proposed coarse-to-fine dataset (C2F-Aff) and training recipe demonstrate a clear performance enhancement.  It is noteworthy that directly employing EVF-SAM to assess results can be highly unsatisfactory (see the first row), which is consistent with the analysis in Introduction and Figure~\ref{fig:difference} that most vision foundation models like SAM lack the capability to recognize affordance verbs and thereby overly segments objects. Encouragingly, after trained on our three consecutive part of C2F-Aff in a stepwise manner, it can be found that the affordance performance has been progressively enhanced. Moreover, as shown in the last row, directly combining all the data can lead to worse result, this may because directly mixing different quality of data and ground-truth format causes the performance degeneration, which is consistent with the observations in tuning LLMs to MLLMs ~\cite{llava,internvl,qwenvl}.
\begin{figure}[h!]
    \centering
    \includegraphics[width=\linewidth]{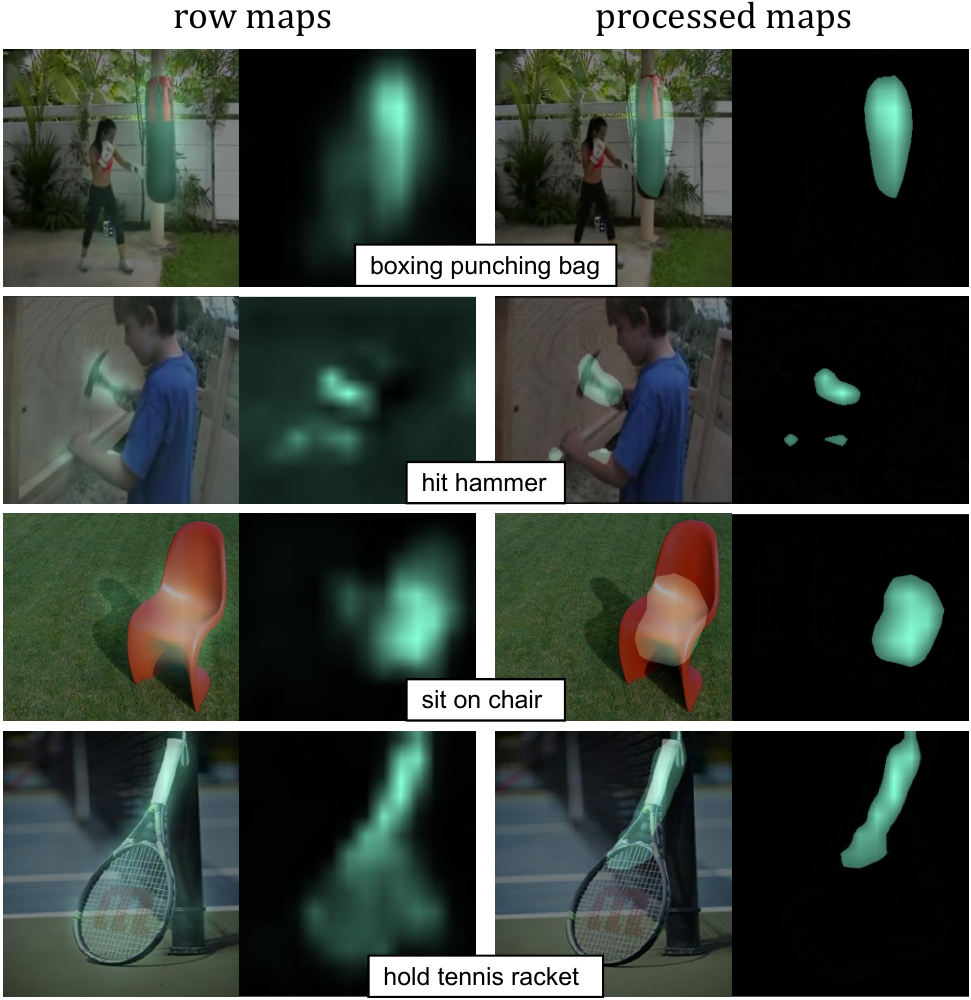}
    \caption{\textbf{Visual comparison between row affordance maps and our processed maps.} It is apparent that our post-processing program successfully mitigate some low thermal regions of the row affordance maps output by LOCATE.}
    \vspace{-0.5em}
    \label{fig:process}
\end{figure}

\noindent\textbf{Design Choices of Affordance-Adaption Module.} We start from EVF-SAM without any structural changes and gradually integrate our methods to
analyze the effect of each proposed design in affordance-adaption module. The results in Tab~\ref{subtab:affcon} reveal that each element consistently delivers notable improvements. In particular, we notice that adding learnable queries benefits a lot, which is in line with findings in other works~\cite{promptlearn1,promptlearn2,promptlearn3} that use prompt learning when adapting a foundation model to downstream tasks.

\noindent\textbf{Effect of Post-Processing Program.} As we display in the first two columns of Figure~\ref{fig:process}, the quality of affordance maps directly output by LOCATE is sometimes terrible, which would result in corruption of the model capability supervised on these maps (see Table~\ref{subtab:postprocess} when $\gamma = 0$). By contrast, after processing with an optimal filtration intensity of $\gamma=0.45$ and a number of 3 (see Table~\ref{subtab:fli_num}), 
most of the irrelevant areas are eliminated, while areas related to the functional part are preserved (see the second two columns of Figure~\ref{fig:process}). This optional design choice also leads to a significant performance improvement across all metrics.

\section{Conclusion}
In this paper, we introduce AffordanceSAM, a new state-of-the-art (SOTA) affordance grounding model. Built upon EVF-SAM with our proposed affordance-adaption module and trained with our systematically curated C2F-Aff dataset in a coarse-to-fine training manner, AffordanceSAM demonstrates superior performance in affordance grounding tasks. The zero-shot affordance grounding evaluation on internet images further demonstrates the generalization capacity of AffordanceSAM. We believe AffordanceSAM can become the new baseline, and the C2F-Aff dataset can become an effective data resource for subsequent studies and offer timely insights into how to leverage and extend SAM-like foundation models to benefit more relevant vision tasks.

\bibliography{aaai2026}
\clearpage
\appendix
\renewcommand{\thesection}{\arabic{section}}
\twocolumn[
  \begin{@twocolumnfalse}
    \section*{\centering Appendix of AffordanceSAM}
  \end{@twocolumnfalse}
]
\section{More Discussion on Related Work}

\noindent\textbf{Affordance Learning for Robotics.} Recently, Numerous methodologies have been developed by researchers to extract and interpret affordance information to enable the robots to grasp in a complex, dynamic environments~\cite{aff_ro1,aff_ro2}. Specifically, several studies~\cite{grasp1,grasp2,glover} leverage affordance to establish correlations between objects, tasks, and manipulation strategies for robotic grasping. Other studies~\cite{real_ro1,real_ro2,real_ro3} focus on deriving affordance knowledge from resources that can be deployed on real robotic systems. In this study, we only focus on the visual affordance grounding task, and our main insight is taking advantage of SAM's generalization ability to compensate for the deficiencies of the previous methods. But we also believe our proposed AffordanceSAM  can greatly improve the robot's ability to recognize and grasp objects and become a default option for robot deployment by other researchers.

\noindent\textbf{Segment anything model and its extension.} The Segment Anything Model (SAM)~\cite{sam1,sam2} is an interactive segmentation framework that produces binary masks in response to various prompt types (points, boxes, and coarse masks). Trained on a comprehensive dataset, SAM exhibits robust generalization capacity for segmenting diverse objects. Building upon SAM, SAM-HQ~\cite{hqsam} addresses the segmentation
quality of SAM by introducing a set of hq-tokens; LISA~\cite{lisa} combines SAM with LLaVA~\cite{llava} to enable a segmentation model with reasoning ability. Our work also starts from SAM, but we focus on converting SAM to be a generalized affordance grounding model with our proposed affordance-adaption module and C2F-Aff dataset and  training recipe.

\section{More Details about Our Datasets}
Here we provide the detailed dataset setup for C2F-Aff training and AGD20K testing of AffordanceSAM. The statistics of each dataset are provided in Table~\ref{tab:datasets_detail}.  Just as shown in this table and Table 1 of the main paper, our C2F-Aff dataset is divided into three parts according to the label format, and we will introduce each part in sequence parts.

\begin{table}[h]
\centering
\caption{Statistics of datasets we used in for our C2F-Aff data. PL.: part-level annotation. Obj.: number of object classes. Aff: number of affordance action classes. Img.: number of images.}
\label{tab:datasets_detail}
\resizebox{0.5\textwidth}{!}{%
\begin{tabular}{l|c|c|c|c|c}
\hline
Dataset & Year  & PL & Obj. & Aff. & Img. \\
\hline
\multicolumn{6}{l}{\emph{Part 1}} \\
PADv2~\cite{pad} & 2021  & $\times$& 103 & 39 & 30,000 \\
RGB-D Part Affordance~\cite{partaff} & 2022  &$\checkmark$ & 37 & 15 & 47,210 \\
Handal~\cite{handal} & 2023  &$\checkmark$ & 212 & 17 & 30,800 \\
\hline
\multicolumn{6}{l}{\emph{Part 2 and Part 3}} \\
AGD20k~\cite{AGD20K} & 2021 & $\checkmark$ & 50 & 36 & 23,816 \\
\hline
\end{tabular}%
}
\end{table}

\begin{table*}[t]
\setlength{\linewidth}{\textwidth}
\setlength{\hsize}{\textwidth}
\centering
\caption{Training hyperparameters of AffordanceSAM in each stage.}
\resizebox{0.6\textwidth}{!}{
\begin{tabular}{lccc}
    \toprule
    \textbf{Name} & \textbf{Stage 1} & \textbf{Stage 2} & \textbf{Stage 3}  \\
    \midrule
    Learning rate & 2e-5 & 2e-5 & 4e-5 \\
    Learning rate scheduler & Cosine decay & Cosine decay & Cosine decay \\
    Epochs & 13 & 13 & 26 \\
    LR warmup epochs & 1 & 2 & 0 \\
    Total batch size & $32$ & $32$ & $32$  \\
    Optimizer & AdamW & AdamW & AdamW \\
    AdamW - $\beta_1$ & 0.9 & 0.9 & 0.9 \\
    AdamW - $\beta_2$  & 0.999 & 0.999 & 0.999 \\
    Drop path & 0.1 & 0.1 & 0.1 \\
    Gradient clip & 3 & 3 & 3 \\
    $\mathcal{\lambda}_{dice}$ & 0.5 &  -- & -- \\
    $\mathcal{\lambda}_{bce}$ & 1 &  -- & -- \\
    Seed & 42 & 42 & 42 \\ 
    \bottomrule
\end{tabular}
}
\label{tabel:affsam_hyperparam}
\end{table*}

\begin{table*}[t]
\setlength{\linewidth}{\textwidth}
\setlength{\hsize}{\textwidth}
\centering
\resizebox{0.6\textwidth}{!}{
\begin{tabular}{lc}
    \toprule
    \textbf{Name} & \textbf{AffordanceSAM} \\
    \midrule
    \multicolumn{2}{l}{\emph{Segmentation Model}} \\
    Encoder$\&$Decoder & SAM-ViT-H~\cite{sam1} \\
    Image input size & 1024$\times$1024 \\
    Patch size  & 16 \\
    Encoder hidden dimension & 1280 \\
    Global attention blocks & [7, 15, 23, 31] \\
    \midrule
    \multicolumn{2}{l}{\emph{Multimodal Model}} \\
    Encoder & BEIT-3-L~\cite{beit3} \\
    Text tokenizer & XLM-Roberta~\cite{tokenizer} \\
    Image input size & 224$\times$224 \\
    Patch size  & 16 \\
    Encoder hidden dimension & 1024 \\
    \bottomrule
\end{tabular}
}
\caption{Model configuration of AffordanceSAM.}
\label{tabel:affsam_hyperparam_model}
\end{table*}
\noindent\textbf{Part 1.} In this part, we use PADv2~\cite{pad}, Handal~\cite{handal}, and RGB-D Part Affordance~\cite{partaff}. All three datasets are composed of different object and affordance categories, and images of the same affordance and category are treated as one class. The ground-truth of three datasets is binary mask, but a slight different is that Handal and RGB-D Part Affordance label the part of the object according to the affordance action, and PADv2 only gives the mask of the whole object. Moreover, as Handal and RGB-D Part Affordance are obtained from continuous frames of videos, which means many of the images of the same object category are very similar. We randomly sampled 5 images for those that belong to a video clip to avoid a potential over-fitting problem. And for PADv2, we filter duplicate images and add the remaining images to our training data. Noticing that we do not split any object when training, because the output form of this stage is completely different from the output used in the final evaluation.

\noindent\textbf{Part 2 and Part 3.} In these two data parts, we use the labeled and unlabeled parts of AGD20K~\cite{AGD20K} and use LOCATE~\cite{locate} to annotate the unlabeled samples. We have already discussed this point in detail in the main paper. Moreover, in order to test the model's performance accurately and fairly, we strictly follow AffordanceLLM~\cite{affordancellm}  to split AGD20K for training and testing our model. The easy split is the original unseen split of AGD20K, which has a lot of similarities between the objects in the train and test sets. The hard split ensures that there is no overlap between the object categories in the training and testing sets. This setting can help to reflect the model's generalization ability when the similarity between the training data and the test data changes. More details can be found in AffordanceLLM's paper and project.

\section{More Details about Implementation}
In this section, we provide specific training hyperparameters in each stage and model configuration of  AffordanceSAM in Table~\ref{tabel:affsam_hyperparam} and Table~\ref{tabel:affsam_hyperparam_model}, respectively. We use the checkpoint after the whole training in first two stages as the initialization of the next stage. 

\section{Details of Each Metrics for Evaluation}
In this section, we explain the metrics (KLD~\cite{kld}, SIM~\cite{sim}, and NSS~\cite{nss}) to evaluate models. 
\vspace{0.75em}

\noindent
$\bullet$ \textbf{K}ullback-\textbf{L}eibler \textbf{D}ivergence (KLD) measures distribution difference between the predicted affordance map ($M$) and the ground truth ($M'$), which is 
\begin{equation}
   \mathrm{KLD}\left ( M,M' \right )=\sum_{i}M'_{i}\log\left ( \epsilon + \frac{M'_{i}}{\epsilon+M_{i}} \right ), \label{eq:no20}
\end{equation}

\noindent
$\bullet$ \textbf{Sim}iliary (SIM) is also called histogram intersection, which measures the intersection between the predicted affordance map ($M$) and the ground truth ($M'$). The final range is from 0 to 1. It is given by

\begin{equation}
   \mathrm{SIM}\left ( M, M' \right )=\sum_{i}\min\left ( M_{i},M'_{i}\right ),\\
\end{equation}
where  $\sum_{i}M_{i}=\sum_{i}M'_{i}=1$.
\vspace{0.75em}

\noindent
$\bullet$ \textbf{N}ormalized \textbf{S}canpath \textbf{S}aliency (\textbf{NSS}) measures the correspondence between the prediction map ($M$) and the ground truth ($M'$). It is given by

\begin{equation}
   \mathrm{NSS}\left ( M,M' \right )=\frac{1}{N}\sum_{i}\hat{M}\times M'_{i}, \label{eq:no22}
\end{equation}

where $N=\sum_{i}M'_{i}$, $\hat{M}=\frac{M-\mu\left ( M \right )}{\sigma\left ( M \right )}$. $\mu\left ( M \right )$ and $\sigma\left ( M \right )$ are the mean and standard deviation, respectively.

\section{Methods for Comparison}
Affordance grounding methods can be mainly summarized into
two main categories: weakly supervised and fully supervised methods.
For weakly supervised models, they do not train on explicit labels of the affordance map. Instead, they are trained on two views (egocentric and exocentric) of the same object.  As for fully supervised models, they are supervised under dense labels. In what follows, we explain the main idea of the baseline methods that we used for the evaluation and comparison.

\noindent\textbf{Cross-View-AG~\cite{AGD20K}} proposes a novel framework to extract invariant affordance from exocentric interactions and transfer it to egocentric views. This includes an Affordance Invariance Mining module to minimize intra-class differences in exocentric images and an Affordance Co-relation Preserving strategy to align correlation matrices between views for affordance perception and localization.

\noindent\textbf{LOCATE~\cite{locate}} first localizes where the interaction happens and identifies interaction regions in exocentric views, then uses a PartSelect module to extract affordance-specific information, and transfers this knowledge to egocentric views for affordance grounding using only image-level labels.

\noindent\textbf{R-Mamba~\cite{r-mamba}} first extracts feature embeddings
from exocentric and egocentric images to construct the hypergraphs. Then, a Hypergraph-guided State Space (HSS) block is introduced to reorganize these local relationships from a global perspective to locate the affordance regions.

\noindent\textbf{PLSP~\cite{PLSP}} introduces a label refining stage, a fine-grained feature alignment process, and a lightweight reasoning module to boost the performance of weakly supervised affordance learning.

\noindent\textbf{AffordanceLLM~\cite{affordancellm}} leverages LLaVA~\cite{llava} with a new mask token similar to LISA~\cite{lisa} and  introduce depth maps as 3D information to build the main pipeline. A light-weight mask decoder trained from scratch is also introduced to produce an affordance map.

\noindent\textbf{OOAL~\cite{ooal}} proposes a vision-language framework that includes CLIP~\cite{clip} and DINOv2~\cite{dinov2}. To boost the alignment between visual features of DINOv2 and affordance text embeddings extracted by CLIP, several light-weight modules like learnable text prompt tokens and a CLS-guided transformer decoder are introduced.

\section{More Details about Ablation Setup}
In this section, we have a detailed illustration of the evaluation setup of each table in our ablation study. The table index here refers to the index in the ablation part of the main paper.

\noindent\textbf{Table 3.} This table presents the effect of our proposed  C2F-Aff dataset and training recipe. The first line of results is obtained by directly using EVF-SAM's checkpoint to evaluate. Then we gradually add the training data, divided by the dataset part, and use the last checkpoint to evaluate. When combining all the data, we add the affordance-adaption module at the beginning of the training, which is slightly different from our default setting.

\noindent\textbf{Table 4.} This table presents the effect of each component in our proposed affordance-adaption module. The learnable queries in the second row do not conduct any cross-attention with the mentioned features but conduct self-attention twice, as we mentioned in Section~\textcolor{red}{3.2} in the main paper. 

\noindent\textbf{Table 5 and Table 6.} These two tables present the effect of the design choise of our proposed post-processing program. In the first row of Table 5, we do not use the 5$\sim$6 steps in Algorithm~\textcolor{red}{1}. And so on for the second and third rows. Noticing that we only vary five values of $\gamma$ results in Table 6, we believe more optional values of $\gamma$ can be further searched; however, given the constraints of time and computational resources, as well as the diminishing returns on additional searches, we have decided not to pursue further searches. This decision does not affect the significance of our contribution.

\section{Failure Cases and Feature Work}
\begin{figure}[t]
    \centering
    \includegraphics[width=0.8\linewidth]{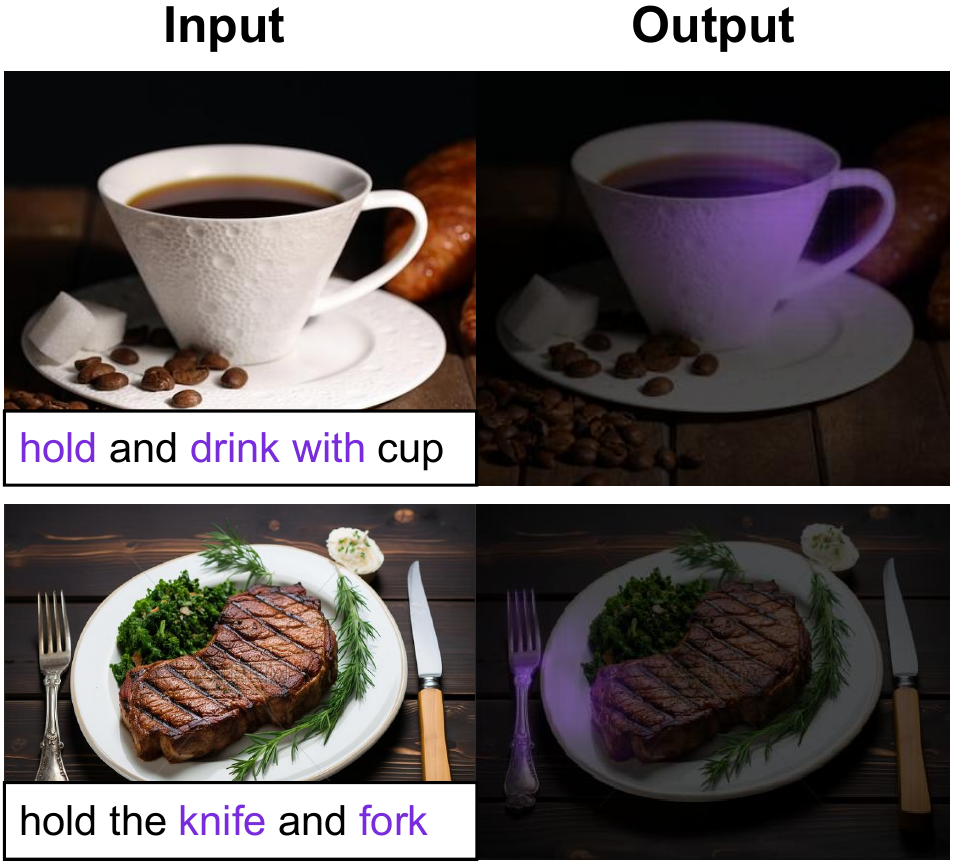}
    \caption{Failure cases when facing multiple objects or multiple affordance actions.}
    \label{fig:failure_case}
\end{figure}
Although our AffordanceSAM achieves state-of-the-art (SOTA) performance under the AGD20K benchmark, and generalizes well beyond the training data, when faced with more complex scenarios (e.g., multi-object affordance), it sometimes cannot get accurate results. As shown in Figure~\ref{fig:failure_case}, first, we find AffordanceSAM fails on completely separating two different functional regions of an object when prompted with multi-actions (see the first row of Figure~\ref{fig:failure_case}). Next, we find that AffordanceSAM can not get an accurate affordance map when two different objects in a picture but are prompted with the same affordance action (see the second row of Figure~\ref{fig:failure_case}). These limitations may be caused by  the fact that every image we use to train AffordanceSAM is a single object format with a single affordance region involved. 

Thus, endowing AffordanceSAM with the capability to deal with more complex scenarios like multiple objects or multiple affordance actions within a single image might be an exciting avenue for future research and development.

\end{document}